\documentclass[letterpaper, 10 pt, conference]{ieeeconf}  % Comment this line out if you need a4paper

\IEEEoverridecommandlockouts                              % This command is only needed if 
\usepackage{graphicx} % for pdf, bitmapped graphics files
\usepackage{amsmath} % assumes amsmath package installed
\usepackage{amssymb}  % assumes amsmath package installed
\usepackage{amsfonts}
\usepackage{xcolor}
\usepackage{epsfig}
\usepackage{graphicx}
\usepackage{subfigure}
\usepackage{epstopdf} % for eps figure using pdflatex
\usepackage[linesnumbered,ruled,vlined]{algorithm2e}
\SetNlSty{bfseries}{\color{black}}{}

\title{\LARGE \bf
Efficient Heuristic Generation for Robot Path Planning\\ with Recurrent Generative Model
}

\author{Zhaoting Li$^{1\dagger}$, Jiankun Wang$^{1\dagger}$ and Max Q.-H. Meng$^{2}$, \emph{Fellow, IEEE}% <-this % stops a space
\thanks{$\dagger$ Equal contribution.}
\thanks{$^{1}$Zhaoting Li and Jiankun Wang are with the Department of Electronic and Electrical Engineering of the Southern University of Science and Technology in Shenzhen, China,
        {\tt\small \{lizt3@mail.,wangjk@\}sustech.edu.cn}}%
\thanks{$^{2}$Max Q.-H. Meng is with the Department of Electronic and Electrical Engineering of the Southern University of Science and Technology in Shenzhen, China, on leave from the Department of Electronic Engineering, The Chinese University of Hong Kong, Hong Kong, and also with the Shenzhen Research Institute of the Chinese University of Hong Kong in Shenzhen, China,(\emph{Corresponding author:} \ {\tt\small max.meng@ieee.org})}%
}

\begin{document}

\maketitle
\thispagestyle{empty} %forces first page to not have a header
%\pagestyle{empty}

%%%%%%%%%%%%%%%%%%%%%%%%%%%%%%%%%%%%%%%%%%%%%%%%%%%%%%%%%%%%%%%%%%%%%%%%%%%%%%%%
\begin{abstract}
Robot path planning is difficult to solve due to the contradiction between optimality of results and complexity of algorithms, even in 2D environments. 
To find an optimal path, the algorithm needs to search all the state space, which costs a lot of computation resource. 
To address this issue, we present a novel recurrent generative model (RGM) which generates efficient heuristic to reduce the search efforts of path planning algorithm. 
This RGM model adopts the framework of general generative adversarial networks (GAN), which consists of a novel generator that can generate heuristic by refining the outputs recurrently and two discriminators that check the connectivity and safety properties of heuristic. 
We test the proposed RGM module in various 2D environments to demonstrate its effectiveness and efficiency.
The results show that the RGM successfully generates appropriate heuristic in both seen and new unseen maps with a high accuracy, demonstrating the good generalization ability of this model. 
We also compare the rapidly-exploring random tree star (RRT*) with generated heuristic and the conventional RRT* in four different maps, showing that the generated heuristic can guide the algorithm to find both initial and optimal solution in a faster and more efficient way.
\end{abstract}

%%%%%%%%%%%%%%%%%%%%%%%%%%%%%%%%%%%%%%%%%%%%%%%%%%%%%%%%%%%%%%%%%%%%%%%%%%%%%%%%
\section{Introduction}

Robot path planning aims to find a collision-free path from the start state to the goal state \cite{choset2005principles}, while satisfying certain constraints such as geometric constraints, and robot kinematic and dynamic constraints. 
Many kinds of algorithms have been proposed to address robot path planning problems, which can be generally classified into three categories.
The grid-based algorithms such as A* \cite{Hart1968Astar} can always find a \emph{resolution optimal} path by searching the discretized space, but they performs badly as the problem scale increases.
The artificial potential field (APF) \cite{khatib1986real} algorithms find a feasible path by following the steepest descent of current potential field.
However, they often end up in a local minimum.
The sampling-based algorithms have gained great success for their capability of efficiently searching the state space, in which two representatives are rapidly-exploring random trees (RRT) \cite{lavalle2001randomized} and probabilistic roadmap (PRM) \cite{kavraki1994probabilistic}.

\begin{figure}[t]
	\centering
	\includegraphics[scale=0.6]{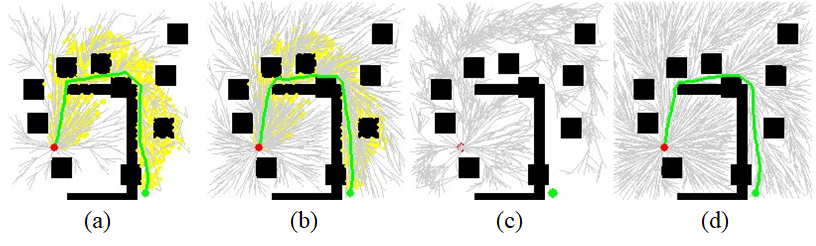}
	\caption{A comparison of conventional RRT* and our proposed heuristic based RRT*. Black area denotes obstacle space and white area denotes free space, respectively. Yellow points denote the generated heuristic from the proposed RGM model. The start and goal states are denoted as red and green circles, respectively. (a)(b) show the planning results of heuristic based RRT* and (c)(d) show the results of conventional RRT*. (a)(c) show the results after 1000 iterations, (b)(d) 3000 iterations.}
	\label{fig:label_illustration_h}
\end{figure}

The sampling-based algorithms have been widely used in our daily life including but not limited to service robot, medical surgery and autonomous driving.
However, the solution from the sampling-based planner is not optimal, resulting in much time cost and energy consuming.
In \cite{karaman2011sampling}, an advanced version of RRT is proposed, namely RRT*, to guarantee an optimal solution as the number of iterations goes to infinity.
But RRT* requires a lot of iterations to converge to the optimal solution, as shown in Fig. \ref{fig:label_illustration_h}.
An effective method is to reduce the sampling cost by biasing the sampling distributions.
A lot of research efforts have been put into the studying of non-uniform sampling, in which especially the deep learning techniques begin to find their effectiveness and generality in robot path planning algorithms.
In \cite{wang2020neural}, Wang \emph{et al.} utilize the convolutional neural network (CNN) to generate the promising region by learning from a lot of successful path planning cases, which serves as a heuristic to guide the sampling process.
Generative models such as generative adversarial nets (GAN) \cite{goodfellow2014generative} and variational autoencoder (VAE) \cite{kingma2013auto} are also popular in learning similar behaviours from prior experiences.
In \cite{YipMPNet2019} \cite{Marco:planning2018}, VAE and conditional VAE (CVAE) techniques are applied to learn sampling distributions.
Although the quality of outputs generated by GANs have seen great improvement recently because of the rapid studying of generator architecture \cite{radford2015unsupervised} \cite{karras2019style} \cite{isola2017image}, loss function \cite{arjovsky2017wasserstein}, and training techniques \cite{mescheder2018training}, there are few researches about the application of GANs on path planning problems.

In this paper, we present a novel recurrent generative model (RGM) to generate efficient heuristic for robot path planning.
RGM follows the general framework of GAN, which consists of two adversarial components: the generator and the discriminator.
The major difference between GAN and RGM is that RGM can incrementally construct the heuristic through the feedback of historical information by combining the recurrent neural network (RNN) \cite{mikolov2011extensions} with the generator. 
With this novel architecture, our proposed RGM exhibits the ability to both fit the training data very well and generalize to new cases which the model has not seen before.
Therefore, when applying RGM to conventional path planning algorithm, the performance can get significant improvement,  as shown in Fig. \ref{fig:label_illustration_h}.

\subsection{Related Work}

Several previous works have presented the effectiveness of neural network based path planning methods.
In \cite{YipMPNet2019}, the motion planning networks (MPNet) is proposed which consists of an encoder network and a path planning network.
%The encoder network adopts the contractive autoencoders (CAE) model.
%During the online path planning process, MPNet predicts the next optimal step given the information of obstacles and the goal state. 
Although MPNet is consistently computationally efficient in all the tested environment, it is unknown how MPNet's performance in complex environments such as ``Bug traps".
The work in \cite{Marco:planning2018} is to learn a non-uniform sampling distribution by a CVAE.
This model generates the nearly optimal distribution by sampling from the CVAE latent space.
But it is difficult to learn because the Gaussian distribution form of the latent space limits its ability to encode all the environment. 
\cite{wang2020neural} presents a framework for generating probability distribution of the optimal path under several constraints such as clearance and step size with a CNN. 
However, the generated distribution may have discontinuous parts in the predicted probability distribution when the environment is complex or the constraints are difficult to satisfy. 
\cite{takahashi2019learning} applies a neural network architecture (U-net \cite{ronneberger2015u}) which is commonly used in semantic segmentation to learn heuristic functions for path planning.
Although the framework belongs to the family of GAN, it only verifies the feasibility of U-net structure in path planning problems of environments which are similar to the training set. 
There is no information about how this model will perform in unseen and complex environments.
\cite{choudhury2018data} proposes a data-driven framework which trains a policy by imitating a clairvoyant oracle planner which employs a backward algorithm. 
This framework can update heuristic during search process by actively inferring structures of the environment. 
However, such a clairvoyant oracle might be infeasible in higher dimensions and their formulation is not appropriate for all learning methods in planning paradigms.
Different from the aforementioned methods, our proposed RGM model can generate efficient heuristic in both seen and unseen 2D environments of various types. The RGM model combines RNN with typical encoder-decoder framework. The experiments demonstrate that the RGM model achieves a high accuracy and exhibits a good generalization ability. 

\subsection{Original Contributions}
The contributions of this paper are threefold. 
First, we proposes a novel recurrent generative model (RGM) to generate efficient heuristic to reduce the sampling efforts. 
Second, we demonstrate the performance of the proposed model on a wide variety of environments, including two types of maps which the model has not seen before. Third, we apply our RGM method to conventional RRT* algorithm, showing that the generated heuristic has the ability to help RRT* find both the initial and optimal solution faster.

The remainder of this paper is organized as follows. 
We first formulate the path planning problem and the quality of heuristic in Section \ref{sec:formulation}. 
Then the details of the proposed RGM model are explained in Section \ref{sec:alg}.
In Section \ref{sec:exp}, we demonstrate the performance of proposed RGM model through a series of simulation experiments. 
At last, we conclude the work of this paper and discuss directions for future work in section \ref{sec:conclude}.

\section{Problem Formulation}
\label{sec:formulation}

The objective of our work is to develop an appropriate generative adversarial network that can generate efficient heuristic to guide the sampling process of conventional path planning algorithms. 
The network should take into account the map information, the start state and goal state information (We refer this as state information for simplicity). 

\subsection{Path Planning Problem}

According to \cite{choset2005principles}, a typical robot path planning problem can be formulated as follows. 
Let ${\mathcal{Q}}$ be the robot configuration space. 
Let ${\mathcal{Q}_{obs}}$ denote the configurations of the robot that collide with obstacles. 
The free space is ${\mathcal{Q}_{free} = {\mathcal{Q}}\backslash{\mathcal{Q}_{obs}} }$.
Define the start state as ${q_{start}}$, the goal state as ${q_{goal}}$, then a feasible path is defined by a continuous function ${p}$ such that ${p:[0,1] \rightarrow \mathcal{Q}}$, where ${p(0)=q_{start}}$, ${p(1)=q_{goal}}$ and ${p(s) \in \mathcal{Q}_{free},\forall s \in [0,1]}$. 

Let ${\mathcal{P}} $ denote the set that includes all feasible paths. 
Given a path planning problem ${(\mathcal{Q}_{free},{q_{start}},{q_{goal}})}$ and a cost function ${c(p)}$, the optimal path is defined as ${p*}$ such that ${c(p^{*}) = min \left\{ c(p): p \in \mathcal{P} \right\} }$.
For sampling-based path planning algorithms, finding an optimal path is a difficult task, which needs a lot of time to converge to the optimal solution.
Herein, we define ${\mathcal{P}^{*}} $ as a set of nearly optimal paths, which satisfy ${(c(p)-c(p^{*}))^{2}< c_{th} ,\forall p \in \mathcal{P}^{*}}$, where ${c_{th}}$ is a positive real number. 
The optimal path ${p^{*}}$ is also included in ${\mathcal{P}^{*}} $.

\begin{figure}[t]
	\centering
	\includegraphics[scale=0.65]{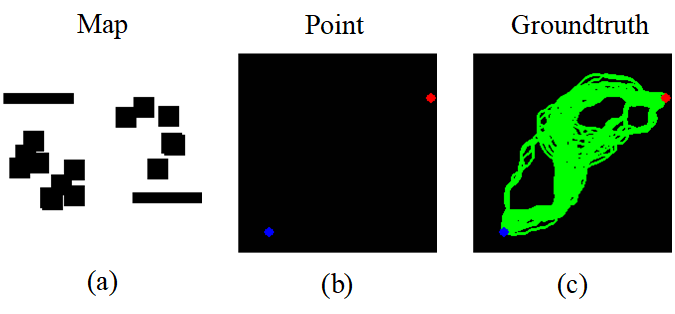}
	\caption{The example of a 2D environment. (a) shows the map information, where the black and white areas denote obstacle space and free space, respectively. (b) shows the information of start state (red point) and goal state (blue point). (c) shows the ground truth.}
	\label{fig:label_data}
\end{figure}

\subsection{Heuristic based Path Planning}

Define ${\mathcal{H}}$ as the heuristic, which is a subspace of  ${\mathcal{Q}_{free}}$ and includes all or part of the optimal path ${p^{*}}$. 

To measure the quality of a heuristic, we define two quality functions ${F^{0}}$ and ${F^{*}}$, both of which take ${({\mathcal{Q}_{free}},\mathcal{H},{q_{start}},{q_{goal}})}$ as input. 
The output of the function ${F^{0}}$ is the number of iterations that sampling-based algorithms take to find a feasible path, while the output of the function ${F^{*}}$ is the number of iterations to find an optimal path. 
We denote ${F^{*}({\mathcal{Q}_{free}},\mathcal{H},{q_{start}},{q_{goal}})}$ as ${F^{*}(\mathcal{H})}$ to simplify the notations.
Therefore, we can obtain the nearly optimal heuristic ${\mathcal{H}^{*}}$ by solving the following equation:

\begin{equation}\label{near_optimal_heuristic}
F^{*}(\mathcal{H}) = F^{*}({\mathcal{P}^{*}}).
\end{equation}

In practice, we found even a non-optimal heuristic can make the planning algorithms achieve a good performance. 
To make this task easier, we define a heuristic ${\mathcal{H}}$ which satisfies the equation \ref{near_optimal_heuristic2} as an efficient heuristic:

\begin{equation}\label{near_optimal_heuristic2}
F^{*}(\mathcal{H})-F^{*}({\mathcal{P}^{*}})  < D_H,
\end{equation}
where ${D_H}$ is a positive threshold value which denotes the maximum allowable deviation from ${\mathcal{P}^{*}}$. 
For example, one feasible heuristic  ${\mathcal{H}}$ is the whole free space ${\mathcal{Q}_{free}}$. 
Obviously, ${\mathcal{Q}_{free}}$ is not an efficient heuristic because it fails to reduce the sampling efforts of the algorithm, which means that 
${F^{*}(\mathcal{H})}={F^{*}({\mathcal{Q}_{free}})}$ and ${F^{*}(\mathcal{Q}_{free}) - F^{*}({{\mathcal{P}^{*}}}) > D_H}$.

In this paper, our goal is to find an efficient heuristic ${\mathcal{H}}$ which satisfies the equation \ref{near_optimal_heuristic2} with a novel recurrent generative model (RGM). 
The main contribution of this paper is to verify the feasibility of neural networks to generate an efficient heuristic in a 2D environment.
One example of ${\mathcal{Q}}$ in our 2D environment is shown in the left of Fig. \ref{fig:label_data}, where the black area denotes obstacle space ${\mathcal{Q}_{obs}}$ and the white area denotes free space $\mathcal{Q}_{free}$.
The start state ${q_{start}}$ and goal state ${q_{goal}}$ of robot are shown in the middle of Fig. \ref{fig:label_data}, where the red point and blue point denote ${q_{start}}$ and ${q_{goal}}$, respectively. 
The set of nearly optimal paths ${\mathcal{P}^{*}} $, which is shown in the right of Fig. \ref{fig:label_data}, is approximated by collecting 50 results of RRT algorithm, given the map and state information. 
Our proposed RGM method takes the map information ${\mathcal{Q}}$, state information ${q_{start}}$ and ${q_{goal}}$ as the input, with the goal to generate a heuristic $\mathcal{H} $ as close to ${\mathcal{P}^{*}} $ as possible. 
For clarity and simplicity, we denote the map information as $m$, the state information as $q$, the nearly optimal path as $p^n$ (also denoted as ground turth) and the generated heuristic as $p^h$.

\begin{figure}[t]
	\centering
	\includegraphics[scale=0.43]{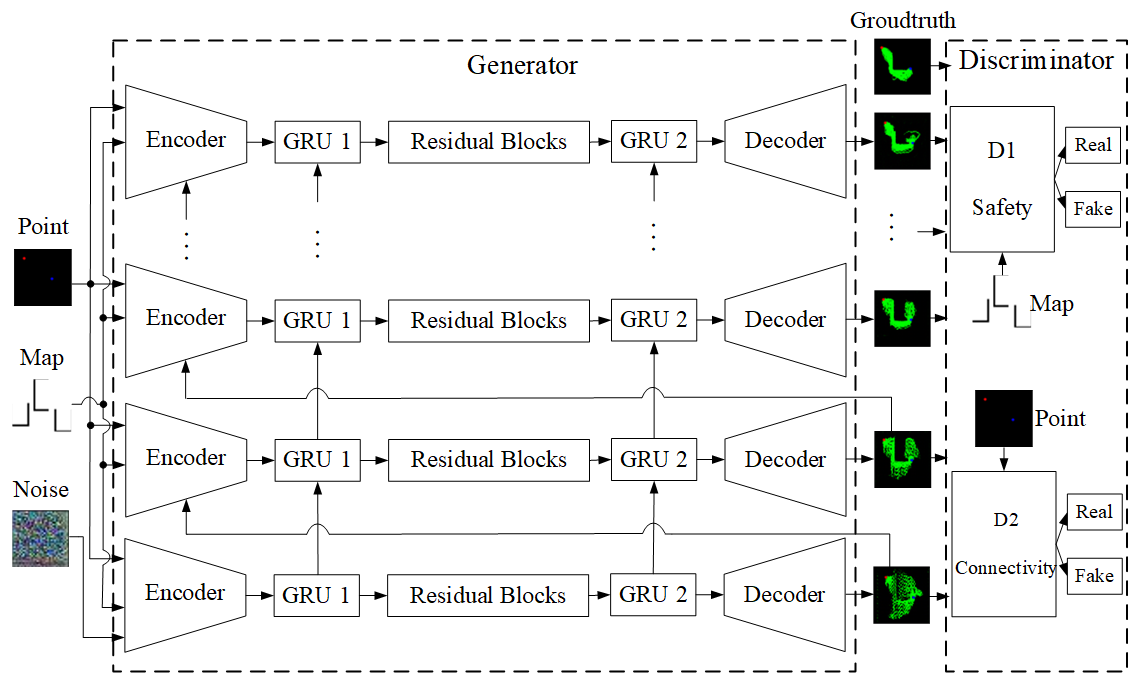}
	\caption{The framework of proposed RGM, which consists of two components: (a) A generator to generate efficient heuristic iteratively; (b) Two discriminators to distinguish generated heuristic from ground truth, from the point of safety and connectivity, respectively. The novel part is the architecture of the generator, which contains two key features: (a) Embed the RNN model into the encoder-decoder framework; (b) Replace the noise input with the output of the previous horizontal process.}
	\label{label_framework}
\end{figure}

\begin{figure*}[t]
	
	\centering
	\includegraphics[scale=0.32]{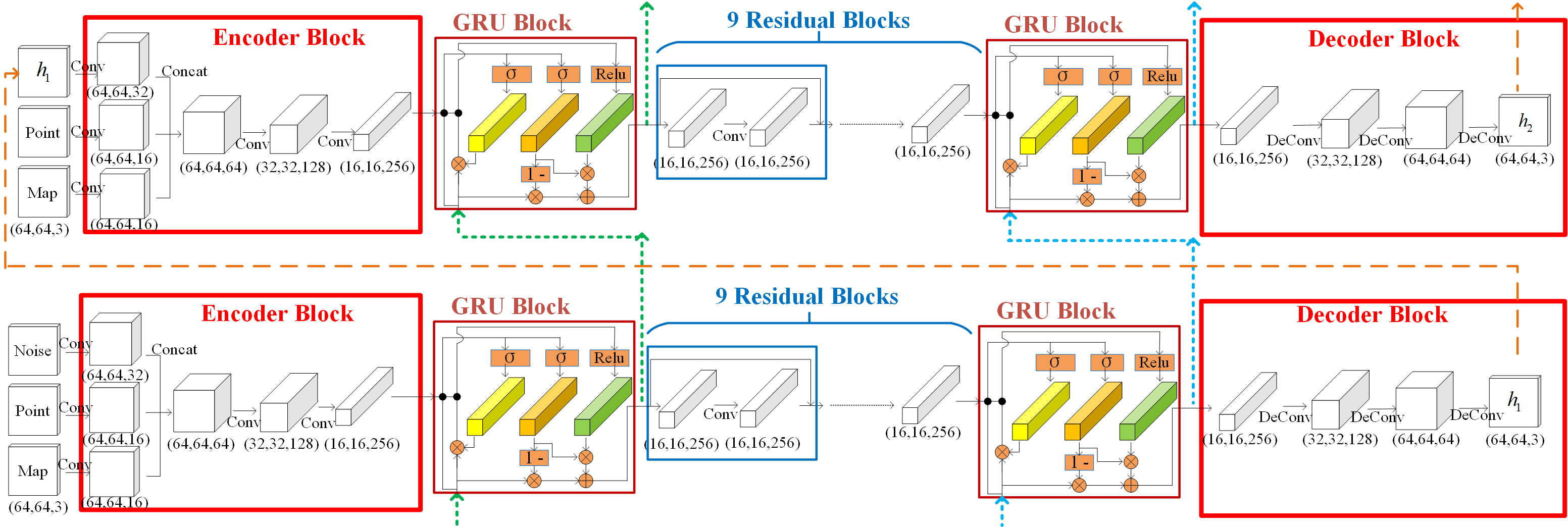}
	\caption{The architecture of our proposed generator, which consists of four key components: (a) An encoder to extract muti-level features from map and state information; (b) GRU blocks to utilize historical information; (c) Residual blocks to improve the generator's ability to generate complicate heuristic; (d) A decoder to generate heuristic from hidden information. Note that apart from part (b), the parameters of $G$ are the same during different iterations.}
	\label{fig:label_generator_framework}
\end{figure*}

\section{Algorithm}
\label{sec:alg}

In this section, we first give a brief introduction to generative adversarial networks. 
Then we present the framework of RGM and illustrate its key components in details.

\subsection{Preliminaries}
We adopt the framework of GAN as the backbone of our model. 
GAN provides a novel way to train a generative model $G$, which is pitted against an adversary: a discriminative model $D$ that learns to determine whether a sample came from the training data or the output distributions of $G$ \cite{goodfellow2014generative}. 
By conditioning the model on additional information $y$, GAN can be extended to a conditional model (cGAN) \cite{mirza2014conditional}.
To learn the generator's distribution $p_g$ over data $x$, the generator $G$ takes noise variables $p_z(z)$ and the conditions $y$ as input and outputs a data space $G(z|y;\theta_g)$. 
Then the discriminator $D$ outputs a single scalar $D(x|y;\theta_d)$ which represents the possibility that $x$ came from the data instead of $p_g$. 
The goal of a conditional GAN is to train $D$ to maximize the $\log{D(x|y)}$ and train $G$ to minimize $\log{(1-D(G(z|y)|y))}$ simultaneously, which can be expressed as:

\begin{eqnarray}
	\operatorname*{min}\limits_{G} \operatorname*{max}\limits_{D}  \mathcal{L}_{cGAN}(G,D) = \mathbb{E}_{x \thicksim p_{data}(x)}[\log{D(x|y)}] \nonumber
	\\ + \mathbb{E}_{z \thicksim p_{z}(z)}[\log{(1-D(G(z|y)|y))}].
\end{eqnarray}

\subsection{Framework Overview}

Herein, both the map information ${m}$ and state information ${q}$ are given by the form of images, as shown in Fig. \ref{fig:label_data}. 
For each pair of map and state information, there is a ground truth information of path ${p^n}$, which is also in a image form. 
The size of these images is $201 \times 201 \times 3$. 
The framework of our neural network is illustrated in Fig. \ref{label_framework}. 
The overall framework is the same as the framework of the typical generative adversarial network, where there are a generator and a discriminator to compete with each other \cite{goodfellow2014generative}.

Different from the typical widely used architecture of the generator such as DCGAN \cite{radford2015unsupervised} and U-net \cite{isola2017image}, our generator combines the RNN with an encoder-decoder framework, which includes an encoder, several residual blocks and a decoder. 
The goal of our model is to train the generator $G$ to learn the heuristic distribution $G(z|m,q;\theta_g)$ over the ground truth $p^n$ under the conditions of state information $q$ and map information $m$, where $z$ is sampled from noise distribution $p_z(z)$.
We denote the heuristic distribution generated by $G$ as $p^h$.
We have two discriminators $D_1$ and $D_2$ which check the safety and connectivity of $p^h$ and $p^n$, respectively.
$D_1$ outputs a single scalar $D_1(h|m;\theta_{d_1})$ which represents the possibility that $h$ comes from the ground truth $p^n$ instead of $p^h$ and $h$ does not collide with map information $m$. 
$D_2$ outputs a single scalar $D_2(h|q;\theta_{d_2})$ which represents the possibility that $h$ comes from the ground truth $p^n$ instead of $p^h$ and $h$ connects state information $q$ without discontinuous parts. 
The generator $G$ tries to produce ``fake" heuristic and deceive the discriminators, while the discriminators try to distinguish the ``fake" heuristic from ``real" heuristic (ground truth).

\subsection{Architecture of Recurrent Generative Model}

The architecture of the generator $G$ is shown in Fig. \ref{fig:label_generator_framework}. 
We resize the images to $(64,64,3)$ for easier handling. 
First, the encoder takes state ${q}$, map ${m}$ and noise ${z}$ as input to extract the important information. 
These inputs are fed into an encoder module using the convolution-BatchNorm-ReLu (CBR) \cite{ioffe2015batch}.
After being concatenated together, the information is fed into the other three encoder modules with the CBR. 
The output of the encoder block has a dimension of $(16,16,256)$. 
Second, this information is fed into a modified Gated Recurrent Unit (GRU) \cite{chung2014empirical} block. 
Different from the typical GRU module, we replace the fully connected layers with the convolutional layer with a ${(3,3)}$ kernel, $1$ padding and $1$ stride step. 
We also replace the tangent function with the CBR when calculating the candidate activation vector.
The output of the GRU block is utilized as the input of residual blocks \cite{he2016deep}, which can improve network's ability to generate heuristic with complex shape. 
Third, the output from residual blocks is stored into another GRU block. 
Then, the output of the GRU block is decoded to generate heuristic that satisfies the condition of ${q}$ and ${m}$. 
Because the information in this process flows horizontally from the encoder to the decoder, while going through the GRU $1$ block, residual blocks and the GRU $2$ block, we define this process as horizontal process ${G_h}$. 
Inspired by drawing a number iteratively instead of at once \cite{gregor2015draw}, the horizontal process ${G_h}$ is executed several times during the recurrent process. 
Define the length of this recurrent process as ${T}$, then the horizontal process at the ${i}$th recurrent process (${i={1,2,..,T}}$) can be denoted as ${G_{hi}}$. 
Define the output of ${G_{hi}}$ as ${h_i}$:

\begin{equation}\label{Output_GAN}
h_i =G_{hi}(z|m,q),z \thicksim p_z^{i}(z)\left\{
\begin{aligned}
 p_z^{i}(z) = h_{i-1},i=2,...,T, \\
 p_z^{i}(z) = p_{z}(z) ,i=1.
\end{aligned}
\right.
\end{equation}

\begin{figure}[t]
	\centering
	\includegraphics[scale=0.45]{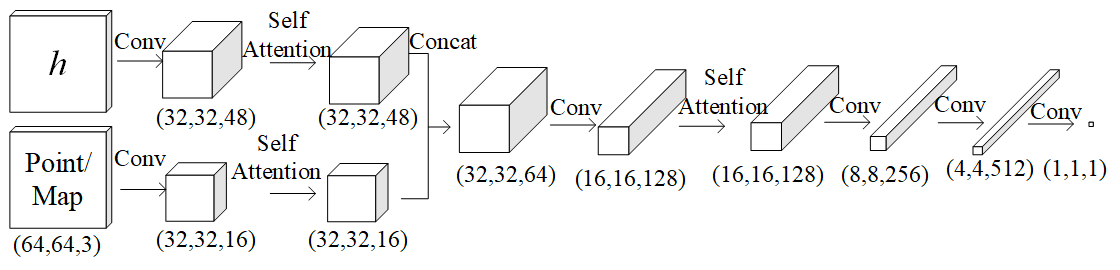}
	\caption{The architecture of the discriminators, which consist of two key components: (a) Convolutional modules with the form of convolution-BatchNorm-LeakyReLU; (b) Self-Attention modules to model long range, multi-level dependencies across different image regions.}
	\label{fig:label_d_framework}
\end{figure}

The architecture of the discriminator $D_1$ and $D_2$ are shown in Fig. \ref{fig:label_d_framework}. 
The ground truth $p^n$ and the corresponding map information $m$ or the state information $q$ are both fed into a convolutional module of the typical convolution-BatchNorm-LeakyReLu (CBLR) form \cite{radford2015unsupervised} and a self-attention module \cite{zhang2019self}, which can learn to find global, long-range relations within internal representations of images. 
Because the quality of heuristic is determined by the whole image, which means that convolutional modules have difficulty to capture the whole information in an image at once, we adopt the self-attention module to make our discriminators distinguish the ``real" and ``fake" heuristic in a global way. 
Then we concatenate the output of these two self-attention layers and feed the information into several convolutional modules with CBLR form and another self-attention layer to output the hidden encoded information with a dimension of $(4,4,512)$. 
At last, we feed the encoded information into a convolutional layer and a sigmoid function to output a score between $[0,1]$.

\subsection{Loss Function}

The objective of training generator $G$ is:
\begin{equation}
	\begin{array}{l}
\displaystyle			\mathcal{L}(G) = \sum_{i=1}^{T} \frac{(i-1)^2}{T^2} \mathbb{E}_{z \thicksim p_{z}(z)}[\log{(1-D_1(G_{hi}(z|m,q)|m))}]
\displaystyle		\\  +\sum_{i=1}^{T} \frac{(i-1)^2}{T^2} \mathbb{E}_{z \thicksim p_{z}^{i}(z)}[\log{(1-D_2(G_{hi}(z|m,q)|q))}],
	\end{array}
\end{equation}
where $G$ tries to minimize this objective. The distribution $p_{z}^{i}(z)$ is defined in equation \ref{Output_GAN}. 
Note that we calculate the weighted average of the scores from discriminators on $h_i, i = 2,3,...,T$. 
This objective can make $G$ pay more attention to its later outputs, allowing $G$ to try different outputs in the beginning. 
During the training process, $G$ learns to generate heuristic which satisfies the standard of $D_1$ and $D_2$. 

The objective of training safety discriminator $D_1$ is:
\begin{eqnarray}
	 \mathcal{L}(D_1) = \mathbb{E}_{h \thicksim p^{n}(h)}[\log{D_{1}(h|m)}] +\nonumber
	\\  \frac{1}{T} \sum_{1}^{T} \mathbb{E}_{z \thicksim p_{z}^{i}(z)}[\log{(1-D_{1}(G_{hi}(z|m,q)|m))}],
\end{eqnarray}
where $D_1$ tries to maximize this objective. 
$D_1$ only takes the map information $m$ and heuristic $h$ as inputs, while checking whether heuristic $h$ collides with the map $m$. 
The objective of training connectivity discriminator $D_2$ is:
\begin{eqnarray}
	\mathcal{L}(D_2) = \mathbb{E}_{h \thicksim p^{n}(h)}[\log{D_{2}(h|q)}] +\nonumber
	\\  \frac{1}{T} \sum_{1}^{T} \mathbb{E}_{z \thicksim p_{z}^{i}(z)}[\log{(1-D_{2}(G_{hi}(z|m,q)|q))}],
\end{eqnarray}
where $D_2$ also tries to maximize this objective. 
The goal of $D_2$ is to check whether heuristic $h$ connects the start state and the goal state. 
The reason we split the discriminator part into two discriminators with different functions is that this framework can help our generator converge in a faster and more stable way. 

\begin{figure}[t]
	\centering
	\includegraphics[scale=0.6]{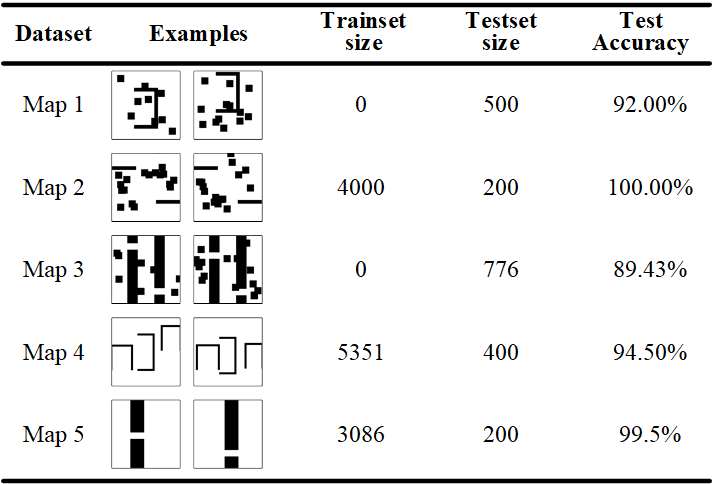}
	\caption{Information about our dataset, consisting of maps which belong to five different types. The test accuracy is also presented. }
	\label{fig:dataset}
\end{figure}

\begin{figure}[!htb]
	
	\centering
	\includegraphics[scale=0.6]{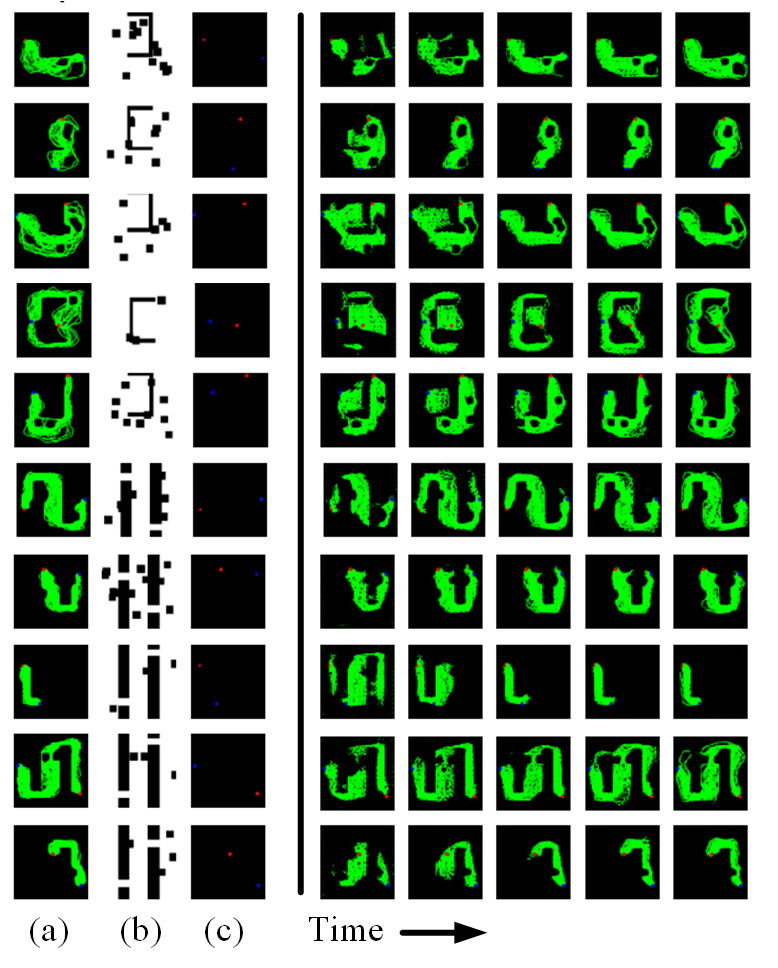}
	\caption{A trained RGM model generating efficient heuristic recurrently given the map information (b) and state information (c). The corresponding groundtruth is in column (a). Note that the RGM model has never seen map type $1$ and $3$ (shown in Fig. \ref{fig:dataset}) during the training process.}
	\label{fig:data_s3}
\end{figure}

\begin{figure*}[t]
	
	\centering
	\includegraphics[scale=0.6]{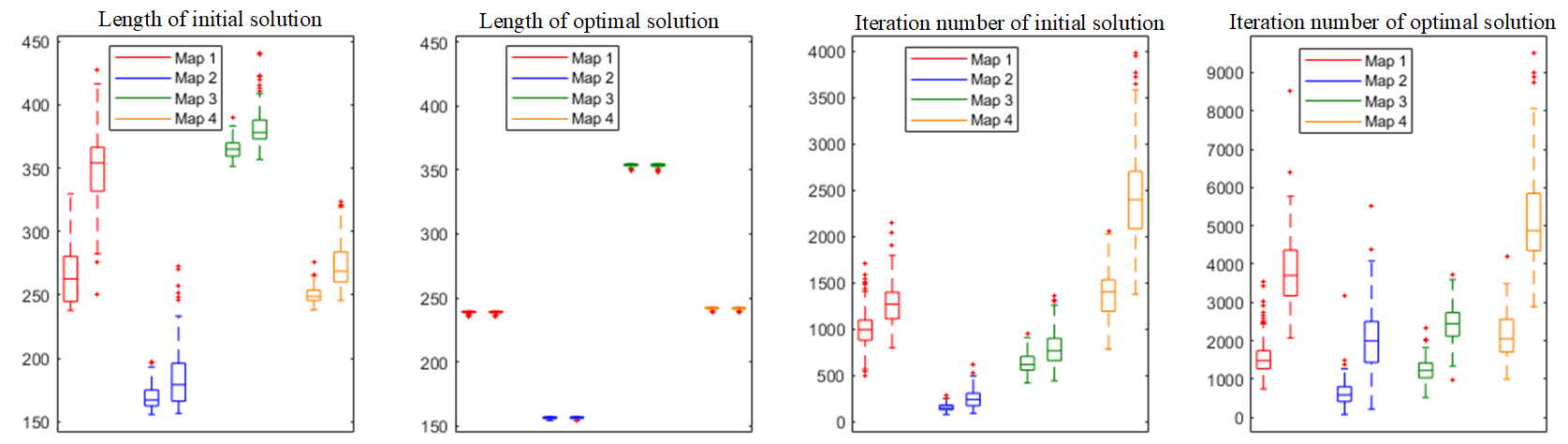}
	\caption{The box pictures for comparison between HRRT* and RRT* on four different maps, which are shown in Fig. \ref{fig:exp_rrtstar}. The data of different maps are presented by different colors. For each map, the left and right part are the data from HRRT* and RRT*, respectively.}
	\label{fig:data_rrtstar}
\end{figure*}

\section{Simulation Experiments}
\label{sec:exp}
In this section, we validate the proposed RGM on five different maps, two of which are never seen by RGM during the training process. 
Then we compare the heuristic based RRT* with conventional RRT* algorithms. 
The simulation settings are as follows.
The proposed RGM model is implemented in PyTorch \cite{paszke2019pytorch}. 
We use Google Colab Pro to train our model and generate heuristic for the data in test set. 
The results in section \ref{sub_exp2} are implemented in PyCharm with Python $3.8$ environment on a Windows $10$ system with a $2.9$GHz CPU and an NVIDA GeForce GTX $1660$ SUPER GPU.

\subsection{Dataset and Implementation Details}
\label{sub_exp1}

We evaluate RGM on various 2D environments which belong to five different types, as shown in the first two columns in Fig. \ref{fig:dataset}. 
The training set is comprised of Map $2$, $4$ and $5$, which includes $200$, $83$ and $112$ different maps, respectively.
For each map we randomly set $20$, $50$ and $20$ pairs of start and end state. 
For each pair of states, we run RRT $50$ times to generate paths in one image. 
Besides, we also set the state information manually to generate data which is relatively rare during the random generation process. 
Therefore, the training set belonging to Map $2$, $4$ and $5$ have $4000$, $5351$ and $3086$ pairs of images, respectively. 
For the test set, the state information is set randomly. The size of our training and test set is shown in the third and fourth column in Fig. \ref{fig:dataset}.

Because the paths in the dataset generated by RRT are not nearly optimal in some cases due to the randomness in RRT algorithm, we adopt one-sided label smoothing technique which replaces $0$ and $1$ targets for discriminators with smoothed values such as $0$ and $0.9$ \cite{salimans2016improved}. 
We also implement the weight clipping technique \cite{arjovsky2017wasserstein} to prevent the neural network from gradient explosion during the training process. 
The test results are presented in Fig. \ref{fig:dataset}, which shows that not only the RGM model can generate efficient heuristic given maps that are similar to maps in training set, and it also has a good generalization ability to deal with the maps that have not been seen before. 
We present examples of heuristic generated by the RGM model, as shown in Fig. \ref{fig:data_s3}.

\subsection{Quantitative Analysis of RRT* with Heuristic}
\label{sub_exp2}

After generating heuristic by the RGM model, we combine the heuristic with RRT* (denoted as HRRT* for simplicity) similar to \cite{Marco:planning2018} and \cite{wang2020neural}: Define the possibility of sampling from heuristic every iteration as $P(i)$. 
If $P(i) > P_h$, then RRT* samples nodes from heuristic, otherwise it samples nodes from the whole map, where $P_h$ is a fixed positive number between $[0,1]$. 
The value of $P_h$ is set manually and needs further research about its impact on the performance of HRRT*. 

\begin{figure}[h]
	\centering
	\includegraphics[scale=0.53]{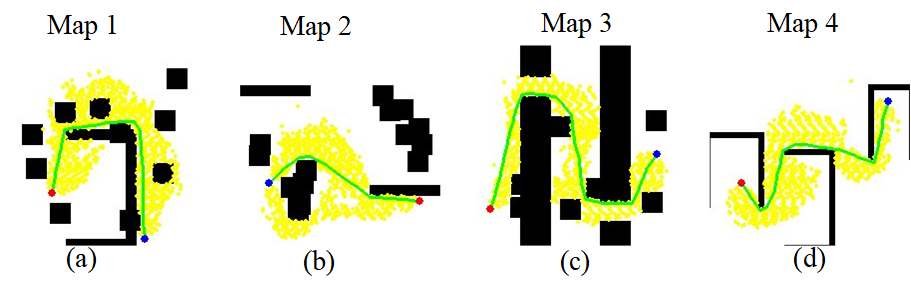}
	\caption{Four maps with different types. Yellow points denote the heuristic.}
	\label{fig:exp_rrtstar}
\end{figure}

We compare this HRRT* with conventional RRT*. 
$P_h$ is set to $0.4$. 
We execute both HRRT* and RRT* on each map $120$ times to get data for statistical comparison, including the length of initial and optimal path and corresponding consumed iterations. 
The experiment results are presented in Fig. \ref{fig:data_rrtstar}. 
Herein, we select four pairs of map and state information, which is shown in Fig. \ref{fig:exp_rrtstar}. 
The discretized heuristic is presented by the yellow points. 
The green paths are generated by HRRT*.
For all the four maps, HRRT* can find an initial path with a shorter length (very close to the optimal path) and fewer iterations compared with RRT*. 
Because the heuristic almost covers the region where the optimal solution exists, HRRT* has more chances to expand its nodes in this region with a constant possibility of sampling from heuristic.
However, RRT* has to sample from the whole map uniformly. 
That is why that HRRT* can also converge to the optimal path with much less iterations, which means that the heuristic generated by RGM satisfies equation \ref{near_optimal_heuristic2} and provide an efficient guidance for the sampling-based path planner.

\section{Conclusions And Future Work}
\label{sec:conclude}
In this paper, we present a novel RGM model which can learn to generate efficient heuristic for robot path planning. 
%RGM consists of a novel generator, which combines the RNN with a typical encoder-decoder framework, and two discriminators which check the safety and connectivity of heuristic respectively. 
The proposed RGM model achieves good performance in environments similar to maps in training set with different state information, and generalizes to unseen environments which have not been seen before.
The generated heuristic can significantly improve the performance of sampling-based path planner.
For the future work, we are constructing a real-world path planning dataset to evaluate and improve the performance of the proposed RGM model.

\addtolength{\textheight}{-8cm}   % This command serves to balance the column lengths

%%%%%%%%%%%%%%%%%%%%%%%%%%%%%%%%%%%%%%%%%%%%%%%%%%%%%%%%%%%%%%%%%%%%%%%%%%%%%%%%
%\section*{ACKNOWLEDGMENT}

%%%%%%%%%%%%%%%%%%%%%%%%%%%%%%%%%%%%%%%%%%%%%%%%%%%%%%%%%%%%%%%%%%%%%%%%%%%%%%%%

\bibliographystyle{IEEEtran}
\bibliography{IEEEabrv,ifacconf}

\end{document}